\documentclass[11pt]{article}

\usepackage[margin=1in]{geometry}
\usepackage[T1]{fontenc}
\usepackage[utf8]{inputenc}
\usepackage{lmodern}
\usepackage{microtype}
\usepackage{amsmath,amssymb,amsfonts}
\usepackage{booktabs}
\usepackage{graphicx}
\usepackage{tabularx}
\usepackage{array}
\usepackage{multirow}
\usepackage{enumitem}
\usepackage{xcolor}
\usepackage{hyperref}
\usepackage[numbers,sort&compress]{natbib}

\hypersetup{colorlinks=true,linkcolor=blue,citecolor=blue,urlcolor=blue}
\newcommand{\ours}{DOA-SORT}

\title{DOA-SORT: Directional Occlusion-Aware Multi-Object Tracking with Distributional Observations}

\author{Hao Wang\\
BDNRC\\
\texttt{wh1090220084@163.co}}
\date{}

\begin{document}
\maketitle

\begin{abstract}
Identity association in multi-object tracking (MOT) is vulnerable to partial
occlusion, truncated detections, and fluctuating confidence scores. Existing
motion-dominant trackers commonly represent occlusion as a scalar penalty.
This treatment misses the directional observation bias caused by occlusion:
left, right, top, and bottom occlusions distort the location and shape of a
detection in different ways. We propose \ours{} (Directional
Occlusion-Aware SORT), an online and training-free tracker that models these
biases explicitly. First, it infers a soft front--back ordering from box
overlap and relative bottom positions, and estimates directional occlusion
coverage and depth. It then constructs a mixture of one clean and four
directional occlusion observation components. The model uses a five-dimensional
observation comprising box center, area, confidence, and aspect ratio, and
adapts observation noise to predicted occlusion and detection confidence. The
directional mixture likelihood is used in high-confidence association,
low-confidence association, and track recovery; ambiguity penalties and local
order-consistency swaps further reduce identity errors among nearby objects.
On the DanceTrack validation split, \ours{} improves HOTA from 63.00 to 66.34,
AssA from 45.10 to 49.57, and IDF1 from 62.19 to 65.28 over OA-SORT with the
same detector and evaluation protocol. The gains are concentrated in
association quality while detection accuracy remains stable. Additional local
evaluations on MOT17 and MOT20 train splits characterize cross-dataset behavior
under the same no-ReID tracking protocol.
\end{abstract}

\section{Introduction}

Multi-object tracking assigns persistent identities to objects detected in a
video. Modern online trackers commonly follow the tracking-by-detection
paradigm: a detector produces bounding boxes for each frame, and a tracker
associates them using motion, appearance, or confidence cues. Although detector
quality has improved substantially, identity switches, fragmented tracks, and
long-term identity maintenance remain difficult in scenes with uniform
appearance and complex motion, such as DanceTrack \citep{sun2022dancetrack}.

During an occlusion, a detector observes only a portion of an object. Its box
center, area, and aspect ratio may therefore show systematic bias, while its
confidence often decreases. Most occlusion-aware association mechanisms reduce
matching weights or relax a motion gate using a scalar occlusion degree. Such a
scalar cannot express which side of an object is occluded or which way its
detection box is expected to move.

We treat occlusion as a structured and directional observation-bias process.
For the same target, occlusion from the left, right, top, and bottom produces
different box deformation statistics. Accordingly, \ours{} represents the
detection under occlusion as a mixture of directional hypotheses and uses this
mixture in data association. The method is built on the score-aware state model
of Hybrid-SORT \citep{yang2024hybridsort} and the occlusion-aware association
framework of OA-SORT \citep{li2026oasort}. It is an online motion model with no
trainable tracking weights and no ReID network.

Our contributions are as follows:
\begin{enumerate}[leftmargin=*,itemsep=2pt]
  \item We introduce a directional occlusion topology and a mixture observation
  model with one clean component and four directional occlusion components.
  \item We extend the geometric observation with detection confidence and use
  confidence- and occlusion-dependent observation noise to represent uncertain
  measurements.
  \item We apply directional mixture likelihoods to primary association,
  low-confidence association, and track recovery, and combine them with
  ambiguity penalties and order-consistency correction.
\end{enumerate}

\section{Related Work}

\subsection{Tracking-by-detection}
SORT \citep{bewley2016sort} combines a Kalman filter with Hungarian matching
for efficient online tracking. DeepSORT \citep{wojke2017simple} augments SORT
with learned appearance features. ByteTrack \citep{zhang2022bytetrack} uses
low-confidence detections in a second association stage, whereas OC-SORT
\citep{cao2023ocsort} improves motion prediction through observation-centric
trajectory correction. BoT-SORT \citep{aharon2022botsort}, Hybrid-SORT
\citep{yang2024hybridsort}, and Deep OC-SORT \citep{maggiolino2023deepocsort}
also use appearance or ReID-enhanced configurations. Our main comparisons use a
no-ReID motion-association protocol, so these appearance-enhanced methods are
not treated as directly comparable baselines.

\subsection{Occlusion-aware association}
OA-SORT \citep{li2026oasort} incorporates occlusion into association scores,
measurement correction, and low-score detection matching. \ours{} retains the
occlusion-aware perspective, but does not collapse occlusion into a single
scalar. Instead, it models conditional observation distributions for different
occlusion directions.

\subsection{Mixture observations and uncertainty}
Gaussian-sum filtering \citep{alspach1972gaussian} and interacting multiple
model filtering \citep{blom1988imm} are classical approaches to multi-hypothesis
state estimation. In contrast, our mixture components are induced by spatial
relations between targets and are used to construct association likelihoods,
without introducing a separate state-switching process.

\section{Method}

\subsection{Overview}
For each frame, \ours{}: (i) predicts active tracks and applies camera-motion
compensation; (ii) estimates an occlusion topology and front--back ordering
from predicted boxes; (iii) builds a clean-plus-directional observation mixture
for each track; (iv) fuses its likelihood with the OA-SORT association cost;
(v) performs high-confidence, low-confidence, and recovery association; (vi)
applies ambiguity penalties and local order-consistency correction; and (vii)
updates matched tracks with the standard Kalman update.

\begin{figure}[t]
  \centering
  \includegraphics[width=\linewidth]{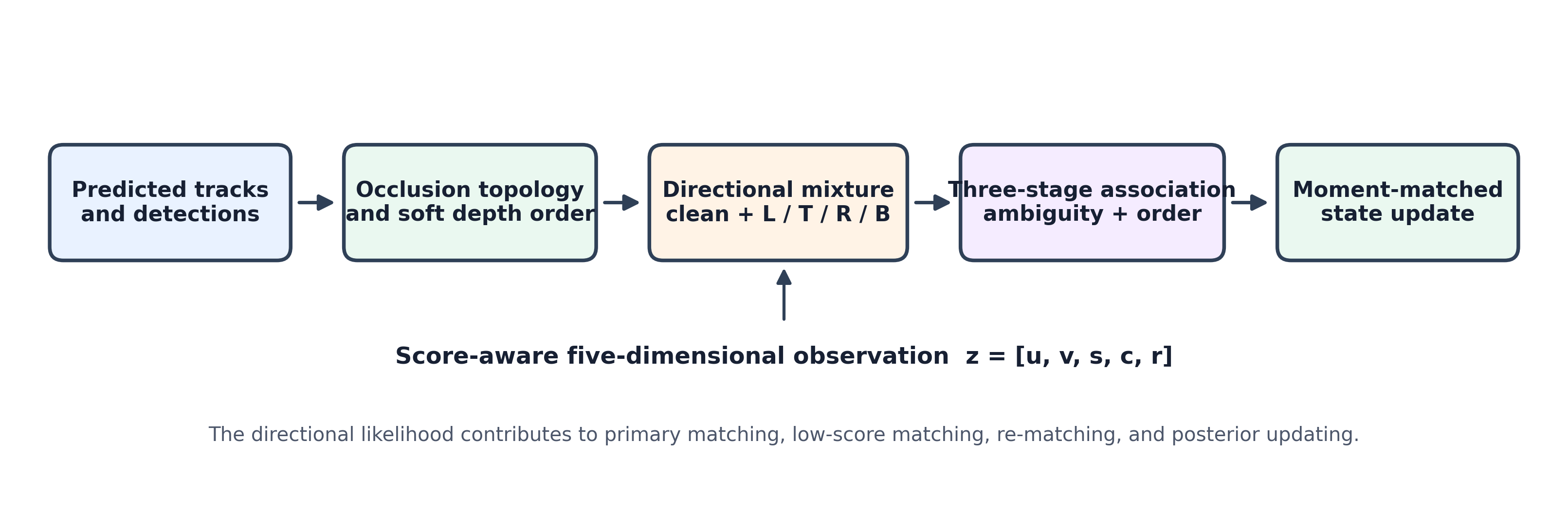}
  \caption{Overview of \ours{}. Predicted tracks and detections induce an
  occlusion topology and soft front--back ordering. A mixture of clean, left,
  top, right, and bottom observation components supplies directional likelihoods
  for multi-stage association. Detection confidence is included in the
  five-dimensional observation.}
  \label{fig:pipeline}
\end{figure}

\subsection{State and observation model}
We use the nine-dimensional Hybrid-SORT state
\begin{equation}
  \mathbf{x}=[u,v,s,c,r,\dot{u},\dot{v},\dot{s},\dot{c}]^T,
  \label{eq:state}
\end{equation}
where $(u,v)$ is the box center, $s$ is box area, $c$ is detection confidence,
and $r$ is aspect ratio. The observation is
\begin{equation}
  \mathbf{z}=[u,v,s,c,r]^T.
  \label{eq:observation}
\end{equation}
Unlike a geometry-only model, \ours{} predicts confidence as a state component
and uses it to control observation uncertainty.

\subsection{Occlusion topology and depth ordering}
Given predicted boxes $\mathcal{B}=\{B_i\}_{i=1}^{N}$, a soft front--back order
is inferred for overlapping boxes from their bottom coordinates:
\begin{equation}
  \tau_i=\max(\lambda(0.5h_i+0.5h_j),\epsilon), \qquad
  p_{ji}=\sigma\!\left(\frac{b_j-b_i}{\tau_i}\right),
  \label{eq:order}
\end{equation}
where $b_i$ is the bottom coordinate of box $i$ and $p_{ji}$ is the probability
that $j$ lies in front of $i$. For each track $i$, we compute directional
coverage and depth vectors
\begin{equation}
 \mathbf{o}_i=[o_i^L,o_i^T,o_i^R,o_i^B], \qquad
 \mathbf{e}_i=[e_i^L,e_i^T,e_i^R,e_i^B].
 \label{eq:occlusion}
\end{equation}
To suppress frame-level noise, directional coverage, directional depth, and
total occlusion are exponentially smoothed, e.g.,
\begin{equation}
 \mathbf{o}_i^t=\alpha\tilde{\mathbf{o}}_i^t+(1-\alpha)\mathbf{o}_i^{t-1},
 \quad
 o_i^t=\alpha\tilde{o}_i^t+(1-\alpha)o_i^{t-1}.
 \label{eq:smoothing}
\end{equation}

\subsection{Directional mixture observation model}
For each track, the observation likelihood is a five-component mixture:
\begin{equation}
 p(\mathbf{z}_t\mid\mathbf{x}_t)=
 \sum_{k=0}^{4}\pi_k\mathcal{N}
 \left(\mathbf{z}_t;H\mathbf{x}_t+\boldsymbol{\mu}_k,
 HPH^T+R+\Sigma_k\right).
 \label{eq:mixture}
\end{equation}
Here $k=0$ is the clean component and $k=1,\ldots,4$ denote left, top, right,
and bottom occlusion. Their weights are
\begin{equation}
 \pi_0=1-o_i, \qquad
 \pi_k=\frac{o_i o_i^{(k)}}{\sum_j o_i^{(j)}}.
 \label{eq:weights}
\end{equation}
For each directional component, $\boldsymbol{\mu}_k$ and $\Sigma_k$ are the
mean and covariance induced by truncating the visible box in measurement space.
We approximate these moments with three-point Gauss--Legendre quadrature,
\begin{equation}
 \boldsymbol{\mu}_k \approx \sum_{q=1}^{3}w_q
 \left[\phi\!\left(B_i^{(k,q)}\right)-\phi(B_i)\right],
 \label{eq:quadrature}
\end{equation}
where $\phi(\cdot)$ maps box coordinates to the measurement space. Directional
occlusion also introduces a downward confidence bias,
\begin{equation}
 \mu_k^{(c)}=-\gamma_o o_i c_i.
 \label{eq:confidencebias}
\end{equation}

\subsection{Adaptive observation noise}
Low-confidence detections under occlusion usually have greater geometric error.
We therefore use
\begin{equation}
 R_t=\operatorname{diag}(\mathbf{r}_t)R\operatorname{diag}(\mathbf{r}_t),
 \label{eq:adaptivenoise}
\end{equation}
with
\begin{equation}
 \mathbf{r}_t=[1+\lambda_o o_i,1+\lambda_o o_i,1+\lambda_o o_i,
 1+\lambda_c(1-c_i)^2,1+\lambda_o o_i].
 \label{eq:noisevector}
\end{equation}
Thus, both geometric and confidence uncertainty rise as predicted occlusion
increases or detection confidence decreases.

\subsection{Multi-stage directional association}
For a detection $d$ and a track $t$, the directional cost is the negative log
mixture likelihood,
\begin{equation}
 C_{dt}^{\mathrm{DOA}}=\mathcal{L}_{\mathrm{mix}}
 (\mathbf{z}_d,\mathbf{x}_t,P_t,R_t),
 \qquad
 C_{dt}=C_{dt}^{\mathrm{base}}+\lambda_{\mathrm{stage}}C_{dt}^{\mathrm{DOA}}.
 \label{eq:association}
\end{equation}
The base cost uses the OAO score in the high-confidence stage,
height-modulated IoU in the low-confidence stage, and similarity between a
track's last observation and the current detection in recovery. Stage-specific
weights make primary association stricter than low-confidence association and
track recovery.

\subsection{Ambiguity penalty and order consistency}
When nearby candidates have similar costs, we penalize small row- and
column-wise margins:
\begin{equation}
 P_{dt}^{\mathrm{amb}}=\lambda_a\exp\left(-\frac{\Delta_{dt}}{\tau_a}\right),
 \label{eq:ambiguity}
\end{equation}
where $\Delta_{dt}$ is the difference between the best and second-best relevant
candidates. After matching, we perform a local order-consistency check. If the
predicted front--back order of two tracks conflicts with that of their matched
detections, the two assignments are swapped only when both swapped edges are
valid, their total cost is within a tolerance of the original cost, and the
ordering conflict is sufficiently confident.

\subsection{Complexity}
With $N_d$ high-confidence detections and $N_t$ active tracks, directional
cost computation evaluates five likelihood components per track--detection
pair. The main additional cost is the mixture computation and occlusion-topology
construction; association remains at most quadratic in the number of tracks and
detections.

\section{Experiments}

\subsection{Dataset, metrics, and implementation}
We evaluate primarily on the DanceTrack validation split, which contains 25
sequences, 25,508 frames, and 225,148 pedestrian annotations. Its uniform
appearance and diverse motion make it well suited to evaluating motion
association and identity preservation under occlusion \citep{sun2022dancetrack}.
We report HOTA, DetA, AssA \citep{luiten2021hota}, MOTA, IDF1, IDSW, and Frag.
CLEAR MOT and IDF1 follow \citet{bernardin2008clear} and \citet{ristani2016performance},
respectively.

All methods use the same YOLO detector, a detection confidence threshold of
0.1, NMS IoU of 0.7, and an input size of $640\times640$. Tracker association
thresholds are controlled by each method's own configuration. The main \ours{}
result uses the stable order-consistency configuration and does not enable the
high-cost joint-enumeration mode.

\subsection{DanceTrack validation results}
Table~\ref{tab:overall} compares ByteTrack, OC-SORT, OA-SORT, and \ours{} on
DanceTrack validation. All results use the same detector, thresholds, and local
TrackEval-based implementation.

\begin{table*}[t]
 \centering
 \caption{Overall results on DanceTrack validation. Higher is better except
 IDSW and Frag.}
 \label{tab:overall}
 \small
 \begin{tabular}{lrrrrrrrr}
 \toprule
 Method & HOTA & DetA & AssA & MOTA & MOTP & IDF1 & IDSW & Frag \\
 \midrule
 ByteTrack & 55.97 & 77.90 & 40.32 & 95.18 & 85.04 & 61.39 & \textbf{1242} & 2340 \\
 OC-SORT & 64.71 & 88.28 & 47.49 & 95.59 & \textbf{93.90} & 63.22 & 1093 & 2139 \\
 OA-SORT & 63.00 & 88.13 & 45.10 & \textbf{96.25} & 93.60 & 62.19 & 1591 & 1994 \\
 \ours{} & \textbf{66.34} & \textbf{88.90} & \textbf{49.57} & 96.20 & 93.68 & \textbf{65.28} & 1497 & \textbf{1971} \\
 \bottomrule
 \end{tabular}
\end{table*}

\begin{figure}[t]
  \centering
  \includegraphics[width=\linewidth]{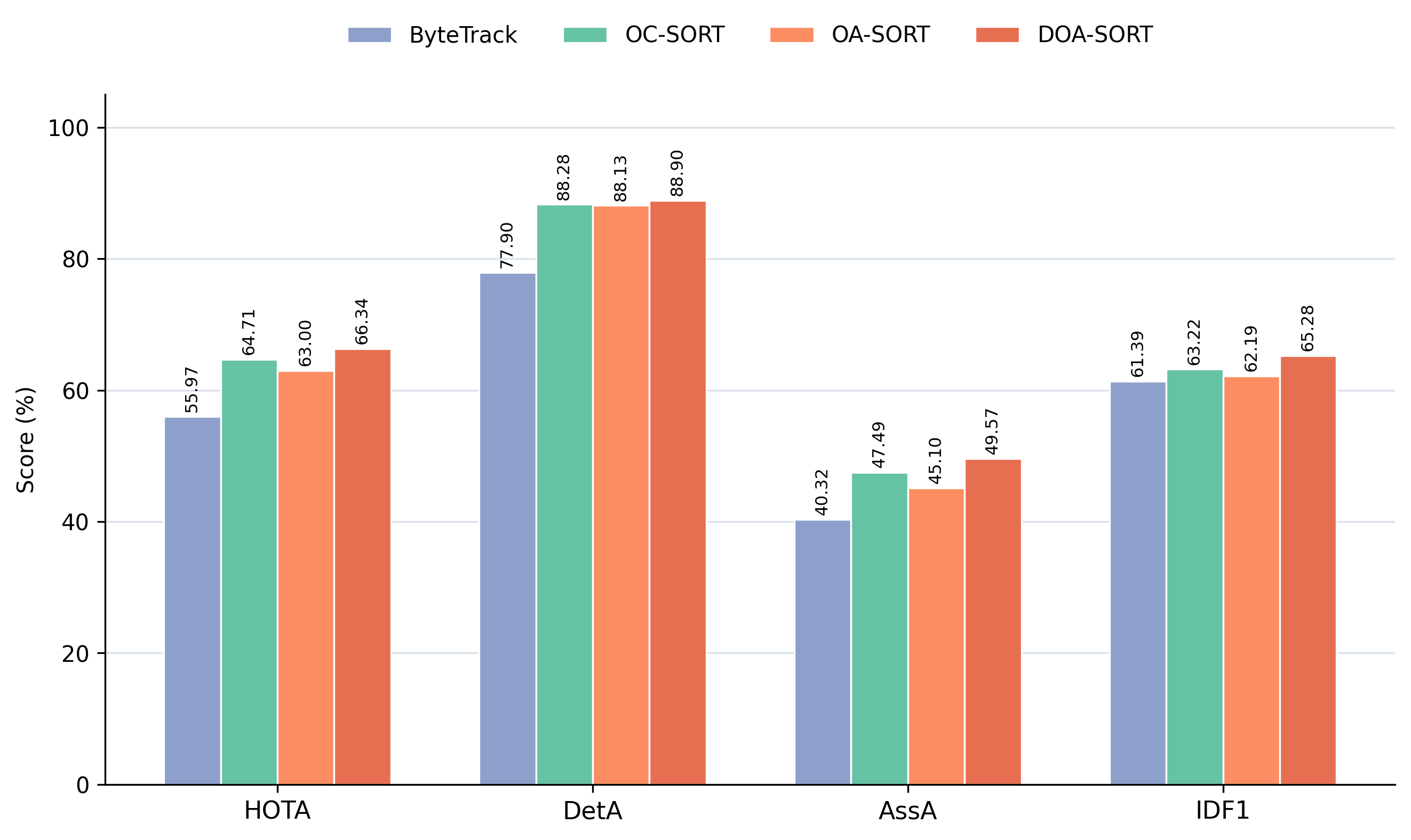}
  \caption{HOTA, DetA, AssA, and IDF1 on DanceTrack validation. Values match
  Table~\ref{tab:overall}.}
  \label{fig:overall}
\end{figure}

Compared with OA-SORT, \ours{} improves HOTA, AssA, and IDF1 by 3.34, 4.47,
and 3.09 points, respectively, while DetA changes by only 0.78 points and MOTA
is essentially unchanged. The improvements therefore arise from association
quality rather than detection coverage. Relative to ByteTrack, \ours{} gains
10.37 HOTA, 11.00 DetA, 9.25 AssA, 8.63 MOTP, and 3.89 IDF1 points. It wins on
HOTA in 23 of 25 sequences and on AssA in 22 of 25 sequences. ByteTrack has
fewer ID switches, but \ours{} has fewer fragments and higher overall HOTA,
AssA, and IDF1. Relative to OC-SORT, \ours{} improves HOTA, DetA, AssA, and
IDF1 by 1.63, 0.63, 2.08, and 2.06 points. Its sequence-level HOTA comparison
is 14 wins versus 11 losses, and the paired test is not statistically
significant, indicating that OC-SORT remains a strong motion-only baseline.

\subsection{Per-sequence and paired analyses}
Table~\ref{tab:sequences} shows representative sequence results. The large
AssA gains on \texttt{dancetrack0047} and \texttt{dancetrack0058} show that
directional observations can mitigate local mismatches in crowded interactions.
The degradation on \texttt{dancetrack0004}, however, shows that fixed mixture
weights can be overly restrictive for some motion patterns.

\begin{table}[t]
 \centering
 \caption{Representative DanceTrack validation sequences.}
 \label{tab:sequences}
 \small
 \begin{tabular}{lrrrrrr}
 \toprule
 Sequence & OA HOTA & DOA HOTA & OA AssA & DOA AssA & OA IDF1 & DOA IDF1 \\
 \midrule
 DT-0014 & 39.80 & \textbf{43.55} & 18.98 & \textbf{22.26} & 35.18 & \textbf{42.50} \\
 DT-0019 & 47.56 & \textbf{49.11} & 25.17 & \textbf{26.97} & 41.36 & \textbf{42.80} \\
 DT-0047 & 61.03 & \textbf{74.59} & 42.03 & \textbf{61.32} & 65.90 & \textbf{77.39} \\
 DT-0058 & 59.74 & \textbf{74.47} & 37.38 & \textbf{57.81} & 58.20 & \textbf{72.56} \\
 DT-0090 & 53.97 & \textbf{56.68} & 33.06 & \textbf{36.60} & 52.82 & \textbf{55.96} \\
 DT-0004 & \textbf{62.75} & 55.66 & \textbf{45.88} & 33.25 & \textbf{66.36} & 48.74 \\
 \bottomrule
 \end{tabular}
\end{table}

\begin{figure}[t]
  \centering
  \includegraphics[width=\linewidth]{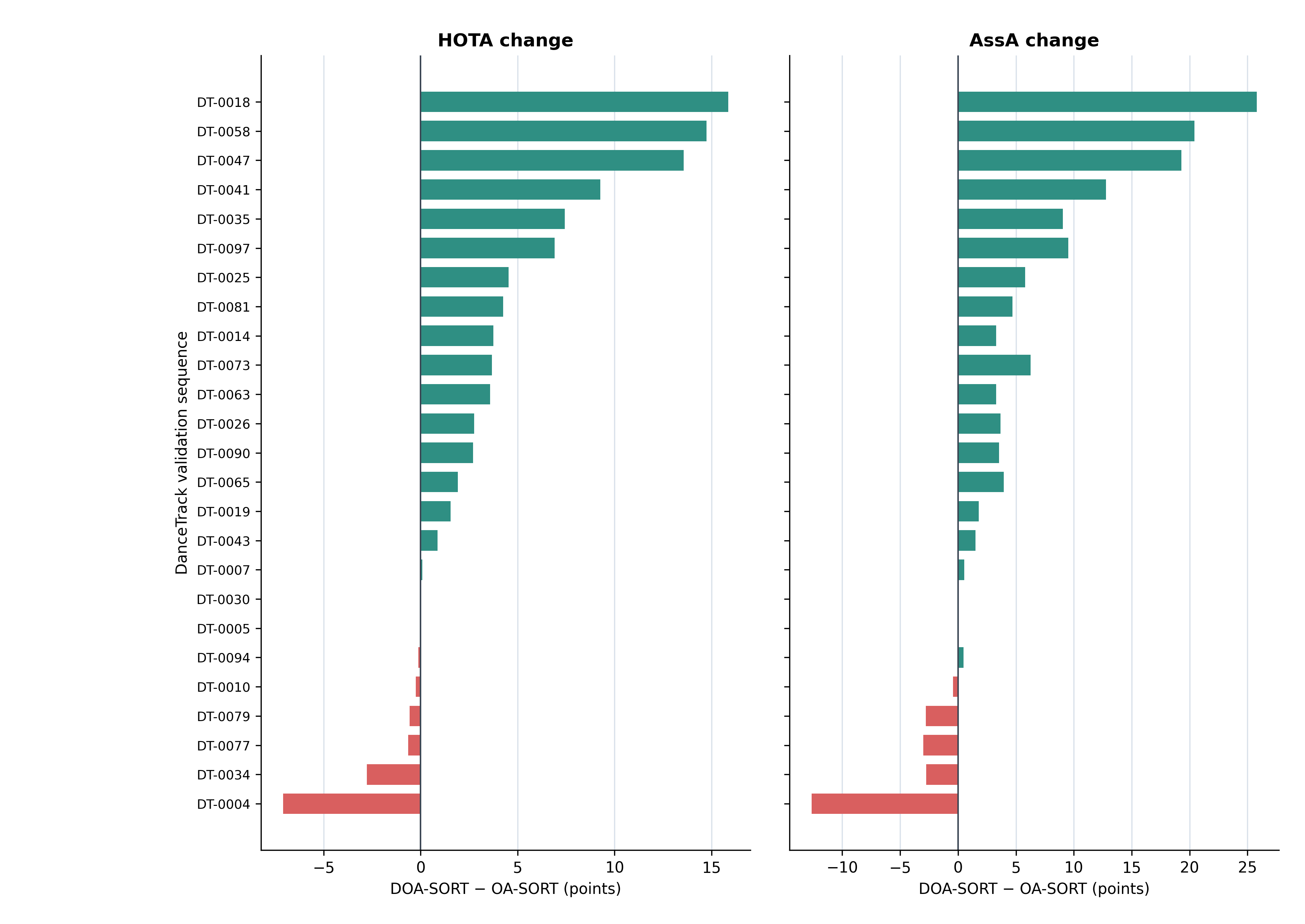}
  \caption{Per-sequence HOTA and AssA differences between \ours{} and
  OA-SORT. Green bars indicate improvements and red bars indicate degradation.}
  \label{fig:deltas}
\end{figure}

The paired Wilcoxon tests in Table~\ref{tab:paired} further show that the gains
over OA-SORT are significant for HOTA, AssA, and IDF1, while the DetA difference
is not significant. The comparisons against ByteTrack are significant for all
reported metrics. Against OC-SORT, the global advantage is consistent but not
statistically significant at the sequence level.

\begin{table}[t]
 \centering
 \caption{Paired sequence-level comparisons.}
 \label{tab:paired}
 \small
 \begin{tabular}{llrrr}
 \toprule
 Comparison & Metric & Wins & Losses & Wilcoxon $p$ \\
 \midrule
 \multirow{4}{*}{DOA vs. OA} & HOTA & 18 & 7 & 0.0031 \\
 & AssA & 19 & 6 & 0.0023 \\
 & IDF1 & 19 & 6 & 0.0081 \\
 & DetA & 16 & 9 & 0.0957 \\
 \midrule
 \multirow{4}{*}{DOA vs. Byte} & HOTA & 23 & 2 & $2.98\!\times\!10^{-7}$ \\
 & DetA & 25 & 0 & $5.96\!\times\!10^{-8}$ \\
 & AssA & 22 & 3 & $2.66\!\times\!10^{-5}$ \\
 & IDF1 & 15 & 10 & 0.0219 \\
 \midrule
 \multirow{4}{*}{DOA vs. OC} & HOTA & 14 & 11 & 0.3123 \\
 & DetA & 15 & 10 & 0.1409 \\
 & AssA & 14 & 11 & 0.5424 \\
 & IDF1 & 14 & 11 & 0.4261 \\
 \bottomrule
 \end{tabular}
\end{table}

\subsection{Ablation study}
Table~\ref{tab:ablation} reports cumulative ablations on DanceTrack validation.
Each setting uses the same detector, thresholds, resolution, and local
TrackEval implementation. A1--A6 add stable modules cumulatively; A7 disables
global motion compensation (GMC) from the stable A6 configuration. MOT17 and
MOT20 official evaluation channels are closed, so this table is a local result
using public DanceTrack validation annotations, not a server score.

\begin{table*}[t]
 \centering
 \caption{Cumulative ablation on DanceTrack validation. Higher is better except
 IDSW and Frag.}
 \label{tab:ablation}
 \resizebox{\textwidth}{!}{%
 \begin{tabular}{llrrrrrrrr}
 \toprule
 ID & Configuration & HOTA & DetA & AssA & MOTA & MOTP & IDF1 & IDSW & Frag \\
 \midrule
 A0 & OA-SORT baseline & 62.78 & 87.97 & 44.88 & 96.18 & 93.58 & 62.69 & 1696 & 2002 \\
 A1 & Directional occlusion mixture & 63.01 & 87.98 & 45.21 & 96.18 & 93.58 & 62.92 & 1691 & 1998 \\
 A2 & Five-dimensional score observation & 63.01 & 87.98 & 45.21 & 96.18 & 93.58 & 62.92 & 1691 & 1998 \\
 A3 & Adaptive observation noise & 63.04 & 87.98 & 45.25 & 96.18 & 93.58 & 62.92 & 1692 & 1998 \\
 A4 & Three-stage directional association & 64.91 & 88.61 & 47.64 & 96.19 & 93.68 & 64.41 & 1590 & 1965 \\
 A5 & Ambiguity penalty & 65.61 & 88.84 & 48.54 & 96.17 & 93.68 & 64.87 & 1659 & 1972 \\
 A6 & Order-consistency correction & 65.89 & 88.96 & 48.88 & 96.18 & 93.67 & 65.30 & 1644 & 1970 \\
 A7 & Stable model without GMC & \textbf{66.34} & 88.90 & \textbf{49.57} & \textbf{96.20} & \textbf{93.68} & 65.28 & \textbf{1497} & \textbf{1971} \\
 \bottomrule
 \end{tabular}}
\end{table*}

A1 improves HOTA, AssA, and IDF1 over A0 by 0.23, 0.33, and 0.24 points. A2
has the same aggregate metrics as A1: confidence enters the state and
likelihood, but does not change final matching decisions under the present
detector and thresholds. A3 yields only 0.03 HOTA and 0.04 AssA, so the global
benefit of adaptive noise is small. A4 is the largest stable association gain,
improving HOTA, DetA, AssA, and IDF1 over A3 by 1.87, 0.62, 2.39, and 1.49
points and reducing IDSW by 102. A5 further improves aggregate association
quality, though IDSW and Frag increase slightly. A6 then gains 0.28 HOTA, 0.35
AssA, and 0.44 IDF1 and reduces IDSW by 15. Finally, disabling GMC in A7 gains
0.45 HOTA and 0.69 AssA relative to A6, while IDF1 is effectively unchanged.
This conclusion applies only to the present DanceTrack validation configuration.

\subsection{Runtime characteristics}
On the same evaluation environment, OA-SORT requires approximately 1688 seconds
for DanceTrack validation and the stable \ours{} configuration requires about
2174 seconds. The extra cost comes mainly from occlusion topology, mixture
likelihoods, and five-dimensional observation processing. These end-to-end
times include both detector and tracker and are not intended as cross-hardware
speed comparisons. The tracker is online and training-free: it has no ReID
network, dataset-specific tracking weights, offline identity features, or
trainable parameters. Its memory is dominated by the detector, runtime tensors,
and per-frame association buffers.

\subsection{Qualitative results}
\begin{figure*}[t]
  \centering
  \includegraphics[width=\textwidth]{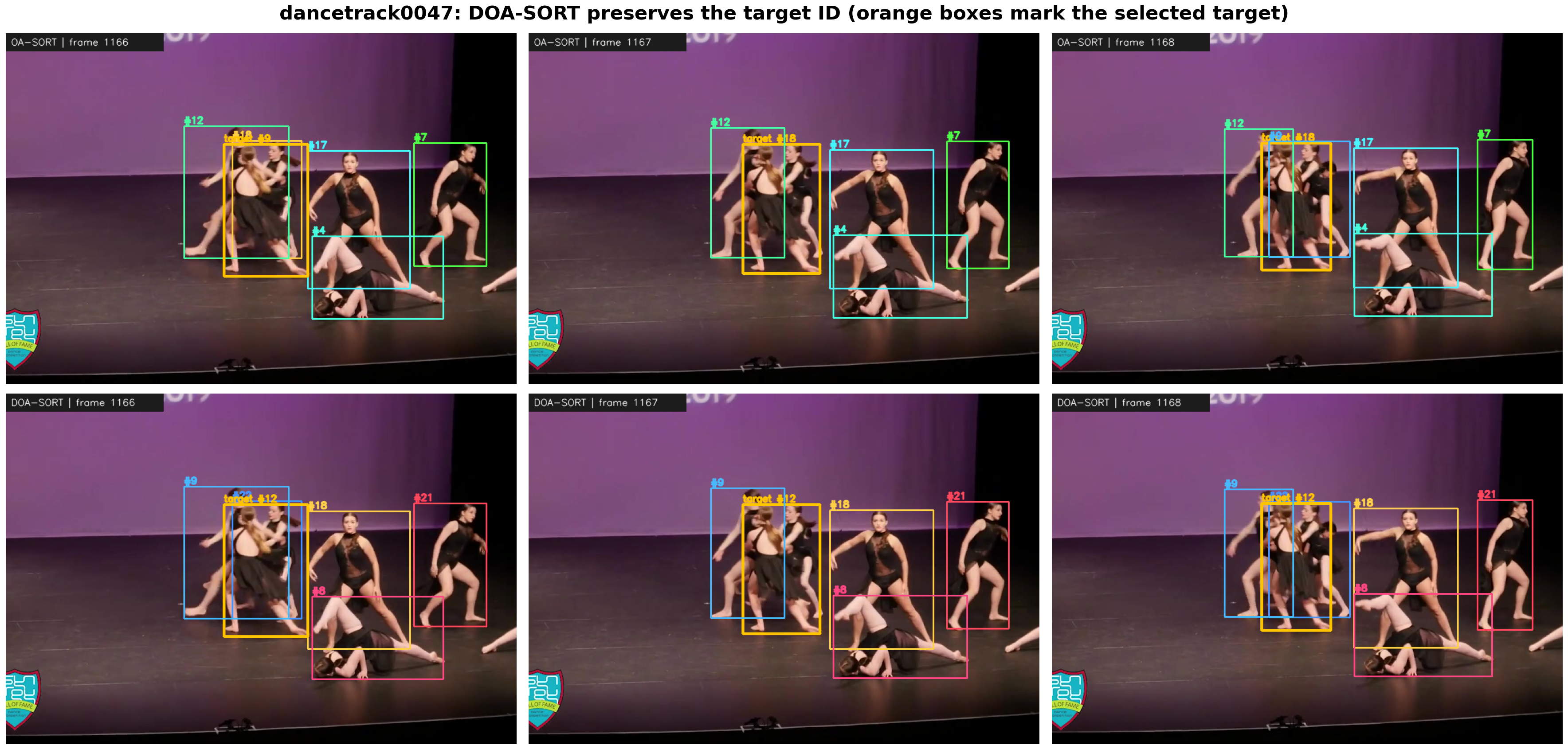}
  \caption{A high-overlap case in \texttt{dancetrack0047}. OA-SORT changes the
  highlighted target identity at the middle frame; \ours{} keeps it consistent.}
  \label{fig:qual47}
\end{figure*}

\begin{figure*}[t]
  \centering
  \includegraphics[width=\textwidth]{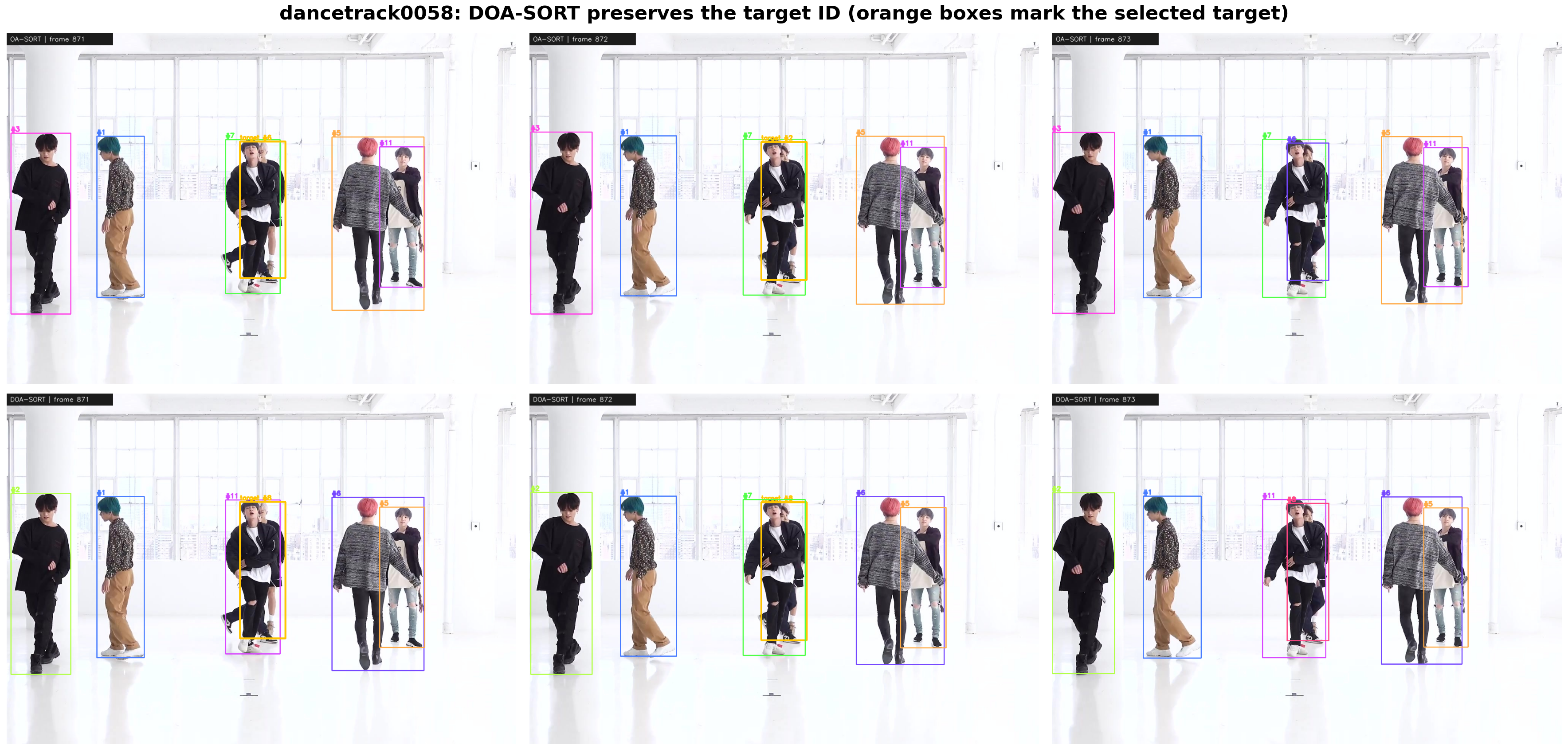}
  \caption{A local-occlusion case in \texttt{dancetrack0058}. \ours{} preserves
  the highlighted identity through the three-frame interaction.}
  \label{fig:qual58}
\end{figure*}

Figures~\ref{fig:qual47} and \ref{fig:qual58} show three consecutive frames.
The orange box marks the same target across frames; other colors and numbers are
tracker-specific identities and should not be compared directly across methods.
In \texttt{dancetrack0047}, OA-SORT changes the target identity from 9 to 18 at
frame 1167, while \ours{} retains identity 12 over frames 1166--1168. The
overlap IoU is 0.835. In \texttt{dancetrack0058}, OA-SORT changes the identity
from 6 to 2 at the central frame, whereas \ours{} preserves identity 8; the
overlap IoU is 0.747.

\subsection{Cross-dataset local evaluations}
To study transfer across density and motion patterns, we additionally evaluate
MOT17 and MOT20 train splits. The MOTChallenge official evaluation channels are
currently closed, so test-server scores cannot be obtained. We instead use
public train-split annotations only for post-hoc metric computation. Ground
truth is not used for detection, association, or state updates. Since \ours{}
is online, training-free, and contains neither a ReID network nor
MOT17/MOT20-specific tracking weights, this setting provides a reproducible
local assessment of tracking behavior, but it is not equivalent to an official
test-server result.

\begin{table*}[t]
 \centering
 \caption{Local evaluation on the MOT17 train split.}
 \label{tab:mot17}
 \small
 \begin{tabular}{lrrrrrrrr}
 \toprule
 Method & HOTA & DetA & AssA & MOTA & MOTP & IDF1 & IDSW & Frag \\
 \midrule
 ByteTrack & \textbf{49.17} & 47.29 & 51.66 & \textbf{56.87} & 78.26 & \textbf{61.76} & \textbf{410} & \textbf{845} \\
 OC-SORT & 48.35 & 44.11 & \textbf{53.45} & 53.51 & \textbf{79.26} & 61.41 & 369 & 1171 \\
 OA-SORT & 47.97 & 47.71 & 48.83 & 55.38 & 78.25 & 59.16 & 717 & 1720 \\
 \ours{} & 49.05 & \textbf{47.95} & 50.75 & 55.16 & 78.19 & 61.25 & 699 & 1830 \\
 \bottomrule
 \end{tabular}
\end{table*}

On MOT17, \ours{} is 0.12 HOTA below ByteTrack while gaining 0.66 DetA. It
improves over OA-SORT by 1.09 HOTA, 1.92 AssA, and 2.08 IDF1, but it does not
consistently exceed ByteTrack or OC-SORT in association. This indicates that the
fixed mixture weights and directional temperature may depend on the target-scale
and motion distributions of DanceTrack.

\begin{table}[t]
 \centering
 \caption{Local evaluation on the MOT20 train split.}
 \label{tab:mot20}
 \small
 \begin{tabular}{lrrrrrr}
 \toprule
 Method & HOTA & DetA & AssA & MOTA & MOTP & IDF1 \\
 \midrule
 ByteTrack & 48.78 & 50.51 & 47.21 & 58.29 & 86.57 & 57.86 \\
 OC-SORT & 44.45 & 46.11 & 42.94 & 51.89 & \textbf{88.26} & 51.13 \\
 OA-SORT & 48.88 & 53.63 & 44.66 & 61.56 & 87.22 & 56.67 \\
 \ours{} & \textbf{51.58} & \textbf{54.21} & \textbf{49.20} & \textbf{62.37} & 87.14 & \textbf{60.73} \\
 \bottomrule
 \end{tabular}
\end{table}

On MOT20, \ours{} improves HOTA, DetA, AssA, MOTA, and IDF1 over OC-SORT by
7.13, 8.09, 6.26, 10.48, and 9.60 points. Relative to OA-SORT, the corresponding
improvements are 2.70, 0.58, 4.54, 0.81, and 4.06 points. These local train-split
results indicate that directional occlusion observations are particularly useful
in dense and heavily occluded scenes, but they must not be interpreted as
official MOTChallenge test scores.

\section{Discussion}
\subsection{Why the gains concentrate on association}
\ours{} does not change the detector or modify its output policy. Consequently,
large DetA gains are not expected. Directional mixture observations act on
matching likelihoods and association decisions, which explains why the main
gains are in AssA, IDF1, and identity continuity.

\subsection{Failure modes and limitations}
The directional topology depends on overlap between predicted boxes; predictions
can drift after long gaps and reduce the accuracy of the topology. Fixed mixture
weights and noise scales may be too strong for some sequences. The current
order-consistency correction is frame-local and cannot recover long fragmented
tracks, and rapid turns or group crossings may violate the linear-motion
assumption. The primary gain is observed on DanceTrack validation. The MOT17
result demonstrates that the present method remains sensitive to target scale,
occlusion density, and motion distribution. Because official MOTChallenge
evaluation channels are closed, official MOT17/MOT20 test-server results cannot
be added; future work should use an active benchmark or an independent test
split and introduce online calibration or sequence-adaptive weighting.

\section{Conclusion}
We presented \ours{}, a directional distributional-observation tracker for
occluded multi-object tracking. It constructs a directional occlusion topology,
uses a clean-plus-four-directional mixture likelihood, and combines
five-dimensional score-aware observations, adaptive uncertainty, multi-stage
association, ambiguity penalties, and order-consistency correction. On
DanceTrack validation, the method improves AssA, IDF1, and HOTA without
materially changing detection metrics. Future work will investigate cross-frame
tracklet association, lost-track re-entry, adaptive weights, and more robust
cross-dataset generalization.

\appendix
\section{Configuration for the Main Result}
\begin{verbatim}
doa_use_direction: true
doa_use_association: true
doa_use_update: false
doa_stage_mode: all
doa_assignment_mode: legacy
doa_temporal_alpha: 0.5
doa_bias_scale: 1.0
doa_gate: 13.28
doa_iou_weight: 0.25
doa_unmatched_cost: 1.25
doa_bottom_temperature: 0.05
doa_primary_weight: 0.35
doa_secondary_weight: 0.25
doa_fallback_weight: 0.40
doa_score_bias_scale: 0.25
doa_score_noise: 0.05
doa_score_uncertainty: 1.0
doa_occlusion_noise_scale: 0.5
doa_assignment_margin: 0.08
doa_ambiguity_weight: 0.10
doa_ambiguity_temperature: 0.08
doa_pairwise_tolerance: 0.05
doa_pairwise_min_confidence: 0.6
doa_order_weight: 0.05
doa_joint_exact_max: 8
\end{verbatim}

\section{Scope of Additional Experiments}
We do not directly compare with ReID methods such as BoT-SORT, Deep OC-SORT,
and Hybrid-SORT because their appearance-feature or ReID configurations are not
equivalent to the no-ReID motion-association protocol in this work. DanceTrack
test-set results are outside the scope of this paper; the reported DanceTrack
results use public validation annotations. Further cross-dataset studies and
parameter-sensitivity analyses are left for future work.

\bibliographystyle{plainnat}
\bibliography{references}
\end{document}